\documentclass[conference]{IEEEtran}
\IEEEoverridecommandlockouts
\usepackage{cite}
\usepackage{amsmath,amssymb,amsfonts}
\usepackage{algorithmic}
\usepackage{graphicx}
\usepackage{textcomp}
\usepackage{xcolor}
\usepackage{booktabs}
\usepackage{amsmath, amssymb, amsfonts}
\usepackage{multirow}
\usepackage{hyperref}
\usepackage{caption}
\def\BibTeX{{\rm B\kern-.05em{\sc i\kern-.025em b}\kern-.08em
    T\kern-.1667em\lower.7ex\hbox{E}\kern-.125emX}}
\begin{document}

\title{S-VGGT: Structure-Aware Subscene Decomposition for Scalable 3D Foundation Models}

\author{
\IEEEauthorblockN{
Xinze Li\IEEEauthorrefmark{1},
Pengxu Chen\IEEEauthorrefmark{2}\IEEEauthorrefmark{1},
Yiyuan Wang\IEEEauthorrefmark{3}\IEEEauthorrefmark{1},
Weifeng Su\IEEEauthorrefmark{1}\IEEEauthorrefmark{4},
Wentao Cheng\IEEEauthorrefmark{1}
}
\IEEEauthorblockA{\IEEEauthorrefmark{1}Beijing Normal-Hong Kong Baptist University}
\IEEEauthorblockA{\IEEEauthorrefmark{2}Jilin University}
\IEEEauthorblockA{\IEEEauthorrefmark{3}Hong Kong Baptist University}\thanks{Corresponding author: Wentao Cheng}
\IEEEauthorblockA{\IEEEauthorrefmark{4}Guangdong Provincial Key Laboratory of Interdisciplinary Research and Application for Data Science}
}

\maketitle

\begin{abstract}
Feed-forward 3D foundation models face a key challenge: the quadratic computational cost introduced by global attention, which severely limits scalability as input length increases. Concurrent acceleration methods, such as token merging, operate at the token level. While they offer local savings, the required nearest-neighbor searches introduce undesirable overhead. Consequently, these techniques fail to tackle the fundamental issue of structural redundancy dominant in dense capture data. In this work, we introduce \textbf{S-VGGT}, a novel approach that addresses redundancy at the structural frame level, drastically shifting the optimization focus. We first leverage the initial features to build a dense scene graph, which characterizes structural scene redundancy and guides the subsequent scene partitioning. Using this graph, we softly assign frames to a small number of subscenes, guaranteeing balanced groups and smooth geometric transitions. The core innovation lies in designing the subscenes to share a common reference frame, establishing a parallel geometric bridge that enables independent and highly efficient processing without explicit geometric alignment. This structural reorganization provides strong intrinsic acceleration by cutting the global attention cost at its source. Crucially, S-VGGT is entirely orthogonal to token-level acceleration methods, allowing the two to be seamlessly combined for compounded speedups without compromising reconstruction fidelity. Code is available at \href{https://github.com/Powertony102/S-VGGT}{Github}.
\end{abstract}

\section{Introduction}
\label{sec:intro}
The advent of 3D foundation models has marked a new phase in multi-view geometry, enabling efficient, optimization-free reconstructions with strong generalization. Early pointmap prediction architectures such as DUSt3R \cite{dust3r} established the foundations for robust and unconstrained feed-forward 3D reconstruction. This progress was further advanced by memory-augmented approaches with persistent or spatial state \cite{cut3r, spann3r}, as well as large-scale reconstruction systems capable of handling thousands of views \cite{must3r, fast3r}. End-to-end feed-forward Structure-from-Motion (SfM) pipelines \cite{light3r, duisterhof2025mast3r} further showed that effective pose and depth estimation is possible with minimal reliance on bundle adjustment.

Building on these advancements, global attention systems such as VGGT \cite{vggt} and other large-scale models \cite{mapanything, pi3} have enabled efficient 3D reconstruction by recovering camera poses, depth maps, and point clouds in a single forward pass. However, a critical bottleneck remains, as the quadratic computational cost incurred by global attention grows rapidly with the input sequence length. This problem is particularly pronounced in dense capture scenarios, where frames exhibit significant visual redundancy, and the information gain from additional frames diminishes. In such cases, the challenge shifts to how we can efficiently extract the geometric details from each frame, ensuring that crucial spatial context is preserved without incurring prohibitive computational costs.

\begin{figure}[t]
    \centering
    \includegraphics[width=1\linewidth]{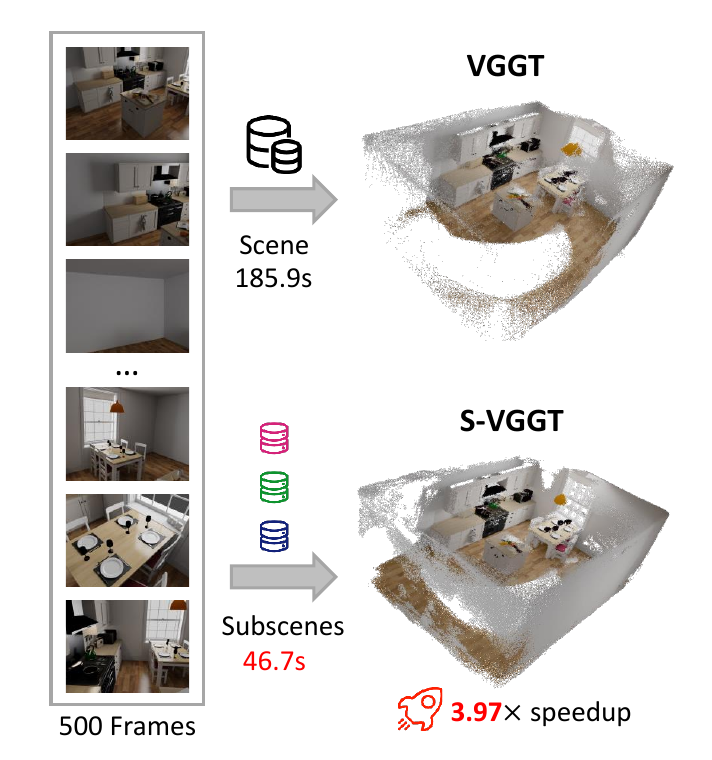}
    \caption{Comparison of VGGT (2.69 FPS) and S-VGGT (10.13 FPS) on a 500-frame scene. S-VGGT achieves a significant speedup by processing subscenes in parallel while maintaining reconstruction quality.}
    \label{fig:teather}
\end{figure}
A prominent line of work focuses on reducing redundancy at the token level. Techniques initially proposed for visual encoders \cite{bolya2023token, fastv} compress local representations by estimating token similarity and merging those judged redundant. Adaptations for 3D geometric models \cite{fastvggt} follow the same principle and have demonstrated meaningful speedups. However, large-scale merging inherently requires computing similarity over a substantial token set, adding nontrivial overhead that partially offsets the acceleration. Furthermore, the fusion of multiple tokens modifies the underlying feature distribution, leading to a representation shift that may degrade geometric precision under high compression ratios.

Our work is motivated by the observation that redundancy in dense sequences manifests primarily at the frame level. In scenarios where neighboring images share significant geometric overlap, the marginal information gain provided by global attention diminishes rapidly, making the quadratic computational cost unjustifiable. This phenomenon mirrors concepts in classical SfM, where strategies such as view graph sparsification \cite{snavely2008skeletal, ni2007out, zhu2018very, shah2018view} and reconstruction via submaps \cite{pan2024global, schonberger2016structure} utilize structural decomposition to tractable subproblems. Our approach capitalizes on this insight: by addressing redundancy at the frame level, we can reduce the effective input sequence length before attention is applied, thereby avoiding the quadratic cost at its source. Crucially, the successful mechanisms used in classical SfM, such as iterative pose refinement and explicit alignment, conflict with the forward only, single pass requirement of modern foundation models. As a result, we require a novel formulation that bridges the gap between structural efficiency and the single pass nature of geometric foundation models.

To realize this vision, we introduce S-VGGT, a framework designed for efficient inference by targeting systematic redundancy. Our core strategy is to decompose the input sequence into a small number of coherent partitions that preserve the original spatial structure while reducing the effective frame count. We leverage the model's intrinsic intermediate features to derive a density-aware affinity score, which captures the inter-frame correlations and guides the sequence decomposition. A key aspect of our approach is assigning a common reference frame to all subscenes. This mechanism allows for independent and parallel processing within a unified coordinate system, eliminating the need for explicit alignment after inference. As a result, we achieve significant reductions in global attention complexity, delivering substantial speedups without sacrificing reconstruction quality. 

We evaluate S-VGGT on several datasets, including ScanNet \cite{dai2017scannet}, NRGBD \cite{nrgbd}, and 7Scenes \cite{7scene}, and demonstrate a significant speed advantage over VGGT. As shown in Fig.~\ref{fig:teather}, S-VGGT achieves substantial acceleration without compromising reconstruction fidelity. Importantly, our sequence optimization is fully orthogonal to token techniques, such as those used in FastVGGT \cite{fastvggt}. This orthogonality enables seamless integration with existing token-based acceleration methods. By combining S-VGGT with techniques like token merging, we achieve even greater speedups, leveraging the strengths of both approaches to reduce computational bottlenecks while preserving reconstruction fidelity.
\section{Method}
\label{sec:method}
\begin{figure*}[htbp]
    \centering
    \includegraphics[width=1.0\linewidth]{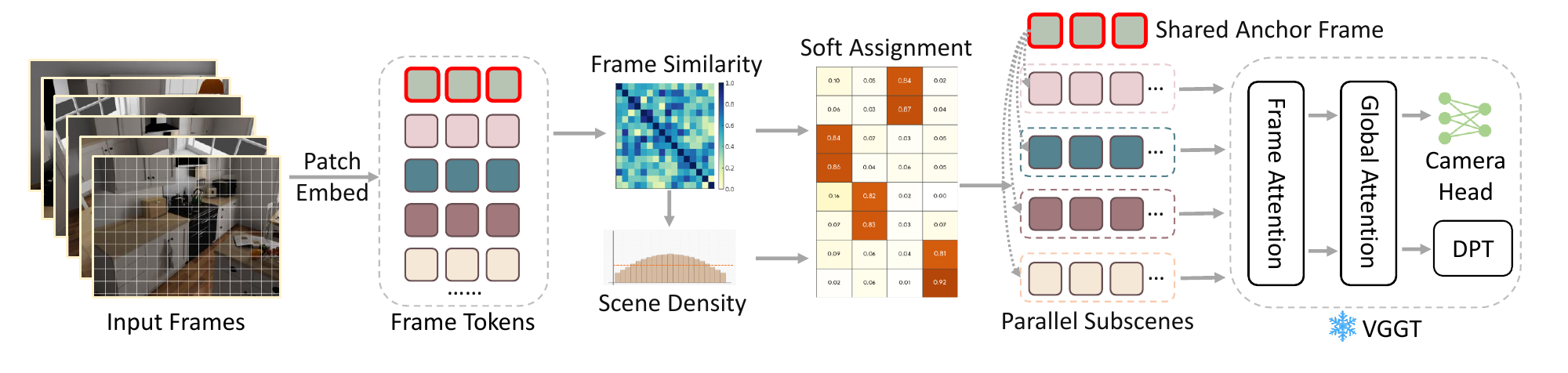}
    \caption{The framework of S-VGGT. The input frames are first embedded into tokens, and frame similarity is calculated to assess redundancy. Frames are then grouped into subscenes via soft assignment, ensuring parallel processing. A shared reference frame across subscenes enables efficient global and frame attention operations, with the model architecture based on VGGT \cite{vggt}.}
    \label{fig:subvggt_framework}
\end{figure*}

In this section, we introduce the key components of our method. First, we provide an overview of the foundational components of VGGT \cite{vggt}, upon which our approach is built. We then detail our strategy for addressing frame-level redundancy through soft partitioning, followed by an explanation of how we evaluate frame similarity and organize frames into coherent subscenes. Next, we describe how the subscenes are processed in parallel and how we ensure consistency across them using anchor frame sharing. Finally, we present the optimization techniques used to achieve efficient grouping and show how they enable significant computational savings without compromising reconstruction quality. The overall framework of our method is illustrated in Fig.~\ref{fig:subvggt_framework}.

\subsection{Preliminaries}

We briefly summarize the relevant components of VGGT \cite{vggt} that our method builds upon.  
Given a sequence of $N$ RGB images $\mathbf{I}=\{\mathbf{I}_s\}_{s=1}^{N}$ with each $\mathbf{I}_s \in \mathbb{R}^{3\times H\times W}$, VGGT first encodes each frame using a DINOv2 \cite{dinov2} backbone:
\[
\mathbf{F}_s = f_{\mathrm{DINO}}(\mathbf{I}_s)\in\mathbb{R}^{P\times C},
\]
producing $P$ patch tokens per frame. Following the original architecture, one camera token and four register tokens are appended, so each frame contains $(P+5)$ tokens in total.

VGGT employs an alternating-attention backbone consisting of $L$ layers.  
Each layer first performs \emph{frame-wise} self-attention independently on all frames, and then applies a \emph{global} cross-frame attention on the concatenation of all tokens:
\[
\mathbf{F} = 
\mathbf{F}_1 \,\Vert\, \mathbf{F}_2 \,\Vert\, \cdots \,\Vert\, \mathbf{F}_N
\in \mathbb{R}^{N(P+5)\times C}.
\]
As a result, every global-attention operation jointly processes $N(P+5)$ tokens, leading to a quadratic computational cost with respect to the sequence length $N$. This global-attention step is the dominant bottleneck when processing dense or long multi-view sequences.

\subsection{Frame Similarity and Scene Density}
\label{subsec:similarity-density}
As in classical incremental SfM pipelines, where a scene graph is built by comparing pairs of images, our framework also begins by evaluating pairwise frame similarity. Although dedicated image retrieval features could be applied here, we find that lightweight descriptors computed directly from the inherited per-frame feature maps in VGGT are sufficient for capturing viewpoint overlap and coarse structural cues. Given the token representation $\mathbf{F}_s$ of frame $s$, we obtain a single $C$-dimensional descriptor by averaging its patch tokens.

Pairwise similarity is computed using cosine similarity,
\[
S_{ij} = \frac{\mathbf{d}_i^\top \mathbf{d}_j}{\|\mathbf{d}_i\|\;\|\mathbf{d}_j\|},
\]
resulting in a similarity matrix $S \in \mathbb{R}^{N\times N}$ that summarizes how strongly each frame is supported by the others. High similarity corresponds to substantial overlap in observed content, whereas low similarity indicates wider viewpoint deviation.

To quantify how densely the frames cover the scene, we count for each frame how many other frames exhibit similarity above a fixed threshold. Averaging this quantity over all frames gives a single density value that reflects the overall redundancy of the input: large values indicate that many frames observe nearly identical content, whereas small values imply substantial viewpoint variation. This density value directly determines the number of subgroups. Instead of relying on manually chosen constants or complex heuristics, we simply clamp the density by a predefined maximum group count $K_{\max}$, using $\min(\text{density},\,K_{\max})$ as the final number of groups. In dense inputs the density is large and thus produces fewer, larger groups, while more diverse inputs naturally yield a larger number of finer groups. This adaptive choice of the group count sets the stage for the grouping mechanism introduced next.

\subsection{Grouping via Soft Assignment}
\label{subsec:grouping}

The objective of this stage is to partition the input frames into $K$ subscenes. Unlike traditional submap-based reconstruction methods, which require costly explicit geometric alignment during final fusion, our approach aims to produce a partitioning that is compatible with efficient feed-forward inference. Concretely, we construct subscenes that satisfy three properties: (1) strong internal connectivity, so that frames within a subscene provide sufficient mutual support for robust local reconstruction; (2) reduced internal redundancy, which lowers the computational overhead; and (3) fidelity to the overall scene geometry, which allows subscenes to share anchor frames and remain implicitly aligned in a common coordinate system.

To achieve the desired grouping behavior, we maintain a soft assignment matrix $\mathbf{A} \in \mathbb{R}^{N \times K}$, where each row encodes a distribution over subscenes and each entry $\mathbf{A}_{sk}$ reflects the degree to which frame $s$ belongs to subscene $k$. This soft formulation remains fully differentiable and GPU-friendly, avoiding the instability and iterative overhead of hard clustering. We optimize $\mathbf{A}$ using three lightweight regularization terms. First, to ensure that each subscene retains meaningful connectivity and remains consistent with the overall scene structure, we favor partitions in which frames share consistent similarity relationships with the rest of the inputs. Each subscene is summarized by a soft group mean $\mathbf{h}_k = \frac{1}{m_k}\sum_{s=1}^{N}\mathbf{A}_{sk}\mathbf{S}_{s:}$ with size $m_k = \sum_{s=1}^{N}\mathbf{A}_{sk}$, and is compared against the global mean $\mathbf{h}_{\mathrm{avg}} = \frac{1}{N}\sum_{s=1}^{N}\mathbf{S}_{s:}$. The coherence loss
\[
\mathcal{L}_{\mathrm{coh}} = \sum_{k=1}^{K} \|\mathbf{h}_k - \mathbf{h}_{\mathrm{avg}}\|_2^{2}
\]
encourages each subscene to preserve global consistency.

To prevent any subscene from becoming disproportionately large, we regularize the soft group sizes \(m_k = \sum_{s=1}^{N}\mathbf{A}_{sk}\) to stay close to the ideal size \(N/K\). This yields a simple balance loss,
\[
\mathcal{L}_{\text{bal}}
=
\sum_{k=1}^{K}
\left(m_k - \frac{N}{K}\right)^2,
\]
which discourages severely unbalanced partitions and helps retain the computational benefits of grouping.

To obtain clear and GPU-friendly discretization, we encourage each row of \(\mathbf{A}\) to form a confident assignment. A lightweight sharpness regularizer,
\[
\mathcal{L}_{\text{sharp}}
=
\sum_{s=1}^{N}\sum_{k=1}^{K}\mathbf{A}_{sk}\,(1-\mathbf{A}_{sk}),
\]
drives the soft assignments toward one-hot vectors without resorting to iterative hard clustering. The full grouping objective,
\[
\mathcal{L}_{\text{group}}
=
\lambda_{\text{coh}}\mathcal{L}_{\text{coh}}
+
\lambda_{\text{bal}}\mathcal{L}_{\text{bal}}
+
\lambda_{\text{sharp}}\mathcal{L}_{\text{sharp}},
\]
is optimized with a small number of gradient descent steps on \(\mathbf{A}\) alone. After optimization, we determine hard assignments by selecting the subscene that maximizes the assignment matrix. Each frame is then associated with a subscene based on this assignment. Each subscene contains a nearly equal number of frames. To ensure they fit within a batch, we perform a lightweight correction by reassigning a few nearby frames based on their similarity. This process remains strictly feed-forward and does not involve any additional model evaluations.

\begin{table*}[t]
    \centering
    \footnotesize
    \captionsetup{singlelinecheck=false}
    \caption{Quantitative results of point cloud reconstruction on the Neural RGB-D\cite{nrgbd} and 7-Scenes\cite{7scene} datasets with 500-frame input sequences. The best results are highlighted in \textbf{bold}, and the second-best results are underlined.}
    \label{exp:quantative-3dreconstruction-1000frames}
    \resizebox{2\columnwidth}{!}{%
    \begin{tabular}{lccccccl ccccccl}
        \toprule
        \multirow{3}{*}{Method} 
        & \multicolumn{7}{c}{NRGBD} 
        & \multicolumn{7}{c}{7 Scenes} \\
        \cmidrule(l{2pt}r{2pt}){2-8} \cmidrule(l{2pt}r{2pt}){9-15}
        & \multicolumn{2}{c}{Acc $\downarrow$} 
        & \multicolumn{2}{c}{Comp $\downarrow$} 
        & \multicolumn{2}{c}{NC $\uparrow$} 
        & \multirow{2}{*}{FPS $\uparrow$}
        & \multicolumn{2}{c}{Acc $\downarrow$} 
        & \multicolumn{2}{c}{Comp $\downarrow$} 
        & \multicolumn{2}{c}{NC $\uparrow$} 
        & \multirow{2}{*}{FPS $\uparrow$} \\
        \cmidrule(l{2pt}r{2pt}){2-7} \cmidrule(l{2pt}r{2pt}){9-14}
        & Mean & Med. & Mean & Med. & Mean & Med. & 
        & Mean & Med. & Mean & Med. & Mean & Med. &  \\ 
        \midrule
        Fast3R 
            & 0.088 & 0.040 & 0.031 & \underline{0.011} & 0.607 & 0.640 & 5.484
            & 0.058 & 0.025 & 0.049 & \underline{0.009} & 0.572 & 0.609 & 5.312 \\
        CUT3R 
            & 0.286 & 0.208 & 0.105 & 0.036 & 0.567 & 0.597 & \textbf{15.342}
            & 0.175 & 0.121 & 0.083 & 0.083 & 0.546 & 0.563 & \textbf{15.435} \\
        Spann3R
            & 0.700 & 0.343 & 0.221 & 0.128 & 0.559 & 0.587 & 7.961
            & 0.379 & 0.242 & 0.163 & 0.080 & 0.534 & 0.548 & 7.895 \\
        \midrule
        VGGT$^*$ 
            & \underline{0.031} & 0.019 & 0.025 & \textbf{0.010} & \textbf{0.642} & \textbf{0.767} & 2.732
            & \underline{0.019} & \textbf{0.009} & \underline{0.028} & 0.010 & \textbf{0.632} & \textbf{0.716} & 2.612 \\
        FastVGGT 
            & \textbf{0.027} & \textbf{0.018} & \underline{0.022} & \textbf{0.010} & \underline{0.638} & \underline{0.764} & 8.092
            & \textbf{0.018} & \textbf{0.009} & 0.029 & 0.010 & \underline{0.625} & \underline{0.702} & 8.011 \\
        Ours 
            & \underline{0.031} & \underline{0.022} & \textbf{0.020} & \textbf{0.010} & 0.622 & 0.717 & \underline{9.934}
            &0.022  & \underline{0.011} & \textbf{0.022} & \textbf{0.008} & 0.622 & 0.697 & \underline{9.425} \\
        \bottomrule
    \end{tabular}
    }
\end{table*}

\subsection{Anchor Frame Sharing}
We partition the sequence into independent subscenes to decouple the attention mechanism, effectively eliminating the quadratic complexity of global attention and allowing the model to ignore irrelevant long-range dependencies. This enables parallel processing of distinct subscenes, helping the system scale to long videos without memory bottlenecks. However, independent processing risks geometric misalignment, as VGGT anchors its 3D coordinate system to the first frame. We address this by introducing Anchor Frame Sharing: by prepending global Frame 0 to each subscene, we ensure that all groups share the same reference point. This simple approach is highly effective in practice. By sharing the anchor frame, all subscenes align in a unified global coordinate system, eliminating the need for complex geometric optimization or rigid alignment, while preserving the efficiency gains from partitioning.

\subsection{Complexity Analysis}
\label{subsec:complexity}

We analyze the computational complexity to demonstrate the intrinsic acceleration of S-VGGT. The bottleneck in the baseline VGGT is the global attention mechanism, which scales quadratically with the sequence length, i.e., $\mathcal{O}((NT)^2)$, where $N$ is the number of frames and $T$ is the number of tokens per frame. By decomposing the sequence into $K$ independent subscenes, S-VGGT reduces the attention cost to $\sum_{k=1}^K \mathcal{O}((NT/K)^2) = \mathcal{O}((NT)^2 / K)$. This yields a theoretical speedup factor of $K$. The overhead introduced by our method—computing frame similarity and soft assignments—scales as $\mathcal{O}(N^2)$ based on frame-level descriptors. Since the number of tokens per frame satisfies $T \gg 1$ (typically $T \approx 1000$), this overhead is negligible compared to the token-level attention cost $\mathcal{O}(N^2 T^2)$. Consequently, S-VGGT achieves substantial reductions in both latency and memory usage.

\section{Experiments}

\begin{table}[t]
    \centering
    \footnotesize
    \captionsetup{singlelinecheck=false}
    \caption{Quantitative results of camera pose estimation and point cloud reconstruction on the ScanNet dataset with input sequences of 1000 images. \textit{OOM} denotes out-of-memory.}
    \label{exp:quantative-scannet}
    \resizebox{\columnwidth}{!}{%
    \begin{tabular}{lcccccc}
        \toprule
        Method & ATE $\downarrow$ & ARE $\downarrow$ & RPE-rot $\downarrow$ & RPE-trans $\downarrow$ & FPS $\uparrow$\\
        \midrule
        Fast3R  & 1.065 & 42.024 & 28.461 & 0.456 & 2.673 \\
        CUT3R   & 1.235 & 56.756 & 0.968 & 0.048 & \textbf{11.725} \\
        Spann3R & \textit{OOM} & \textit{OOM} & \textit{OOM} & \textit{OOM} & \textit{OOM} \\
        \midrule
        VGGT$^*$   & 0.190 & 4.351 & 0.864 & 0.038 & 1.458 \\
        FastVGGT   & \underline{0.162} & \underline{3.805} & \textbf{0.656} & \textbf{0.030} & 5.200 \\
        Ours       & \textbf{0.145} & \textbf{3.576} & \underline{0.665} & \underline{0.053} & \underline{5.699} \\
        \bottomrule
    \end{tabular}
    }
\end{table}
\subsection{Experimental Setup}
\label{subsec:setup}

\paragraph{Datasets and Metrics}
We evaluate our framework on three widely used benchmarks to assess performance across different scene types and trajectory lengths.
ScanNet~\cite{dai2017scannet} serves as the primary benchmark for large-scale camera pose estimation. Following standard protocols~\cite{vggt, dust3r}, we report Absolute Trajectory Error (ATE), Absolute Rotation Error (ARE), and Relative Pose Errors (RPE) on unseen trajectories.
For dense reconstruction quality, we utilize Neural RGB-D~\cite{nrgbd} and 7-Scenes~\cite{7scene}. Here, we report standard geometric metrics: Accuracy (Acc), Completeness (Comp), and Normal Consistency (NC).
To rigorously test scalability and efficiency, we construct long-sequence inputs (500--1000 frames) for all evaluations, challenging the models' ability to handle dense structural redundancy.

\paragraph{Baselines}
We primarily compare S-VGGT against VGGT~\cite{vggt}, the state-of-the-art global-attention foundation model. Specifically, we use the VRAM-efficient variant, denoted as VGGT$^*$, to enable fair comparison on long sequences within memory limits. We also compare against FastVGGT~\cite{fastvggt}, a representative token-level acceleration method.
For broader context, we include recent feed-forward reconstruction baselines derived from DUSt3R~\cite{dust3r}, including Fast3R~\cite{fast3r} (parallel multi-view), Spann3R~\cite{spann3r} (spatial memory), and CUT3R~\cite{cut3r} (recurrent state), to highlight the robustness of our approach in long-sequence settings.

\paragraph{Implementation Details}
All experiments are conducted on a single NVIDIA A100 GPU using \texttt{bfloat16} precision to optimize memory efficiency. For S-VGGT, unless otherwise specified, we employ a maximum subscene count of $K_{\max}=8$ for standard sequences, scaling proportionally for longer inputs. The grouping module performs a lightweight inference-time optimization (typically 10 iterations) to refine soft assignments. Crucially, our framework requires no fine-tuning of the pre-trained VGGT weights; all results are obtained in a strictly zero-shot manner.

\subsection{3D Reconstruction}
\label{subsec:reconstruction}

We evaluate the 3D reconstruction capabilities of S-VGGT on long-sequence inputs from the Neural RGB-D and 7-Scenes datasets. As summarized in Table~\ref{exp:quantative-3dreconstruction-1000frames}, our framework demonstrates a dominant efficiency advantage while preserving high geometric fidelity. In terms of runtime performance, S-VGGT achieves an inference speed of approximately 10 FPS on Neural RGB-D, representing a nearly $3.6\times$ speedup over the VGGT$^*$ baseline (2.73 FPS) and consistently outperforming the token-level acceleration of FastVGGT (8.09 FPS). Similar gains are observed on 7-Scenes, confirming that our frame-level structural partitioning effectively circumvents the quadratic computational cost associated with dense inputs.

Crucially, this acceleration is achieved without compromising reconstruction quality. On both benchmarks, S-VGGT maintains accuracy and completeness metrics comparable to the full-attention baseline. For instance, on 7-Scenes, our method yields an accuracy of 0.022, closely tracking the baseline performance and significantly surpassing DUSt3R-based variants such as Fast3R and CUT3R. This suggests that our density-aware subscenes successfully capture the essential geometric context required for robust depth and point map prediction. Furthermore, unlike methods that struggle with global consistency over long trajectories, our approach maintains high normal consistency, validating the effectiveness of the anchor frame sharing mechanism in preserving a unified coordinate system across parallel subscenes.

\begin{figure}[t]
    \centering
    \includegraphics[width=1.01\linewidth]{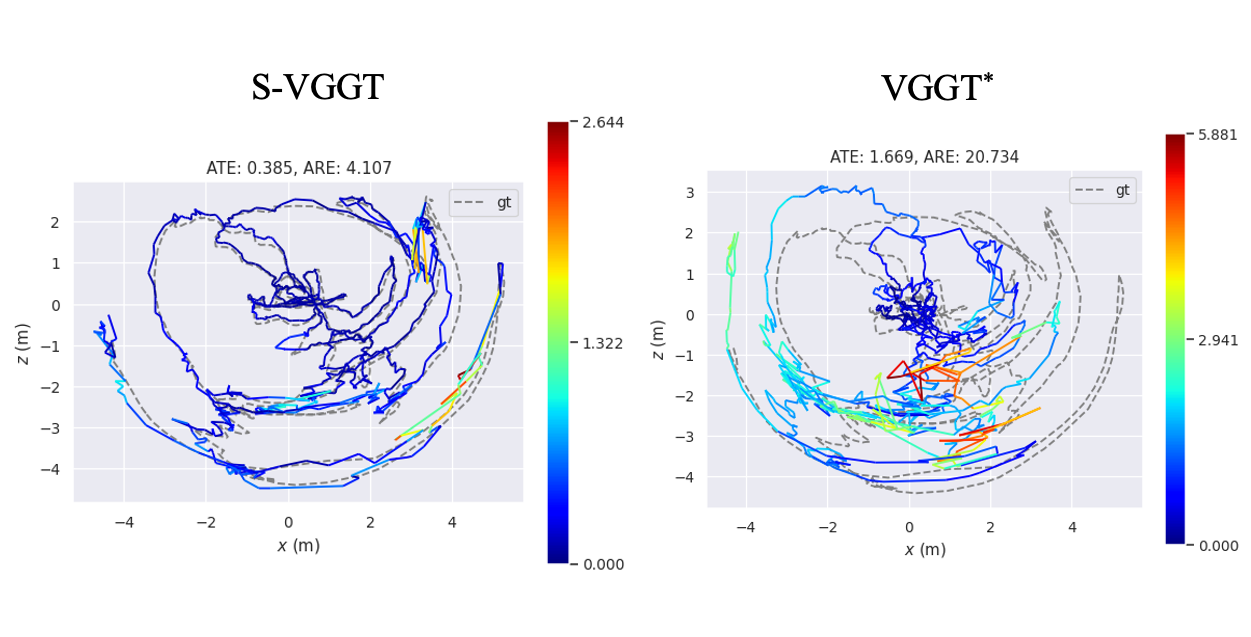}
    \caption{Qualitative comparison of camera pose estimation performance between S-VGGT and VGGT$^*$. }
    \label{fig:camera-pose-estimation}
\end{figure}
\subsection{Camera Pose Estimation}
We evaluate the camera pose estimation performance on unseen trajectories drawn from a uniformly sampled subset of ScanNet~\cite{dai2017scannet}. All experiments are conducted on challenging input sequences of length 1000 images, which demonstrate the crucial efficiency and robustness advantages of our method under long-sequence configurations. As shown in Table~\ref{exp:quantative-scannet}, our framework excels in both accuracy and runtime efficiency. Specifically, S-VGGT achieves the best absolute pose accuracy, with an ATE of 0.145, significantly outperforming the original VGGT$^*$ baseline (ATE 0.190) and even the token-accelerated FastVGGT. This result is particularly notable, as the structural partitioning in our approach, which intentionally limits the scope of attention, unexpectedly improves geometric accuracy.

In terms of efficiency, S-VGGT achieves a remarkable $3.9\times$ speedup over VGGT$^*$ . While methods like CUT3R achieve high raw FPS, their absolute pose estimation performance suffers from accumulated error over long sequences. Similarly, Spann3R, which relies on spatial memory, fails entirely due to memory limitations. The key to our success lies in addressing the accumulated error introduced by global attention in previous VGGT variants, which arises from noisy long-range correlations as the frame count increases. By partitioning the sequence into subscenes, we effectively constrain attention to density-consistent regions, filtering out irrelevant correlations. This design allows S-VGGT to achieve superior prediction quality while maintaining scalability for large, dense datasets. A qualitative comparison of camera pose estimation performance between S-VGGT and VGGT$^*$ is shown in Fig.~\ref{fig:camera-pose-estimation}.

\subsection{Complementarity with Token-Level Acceleration}
\label{subsec:orthogonality}
\begin{figure}[htbp]
    \centering
    \includegraphics[width=1.0\linewidth]{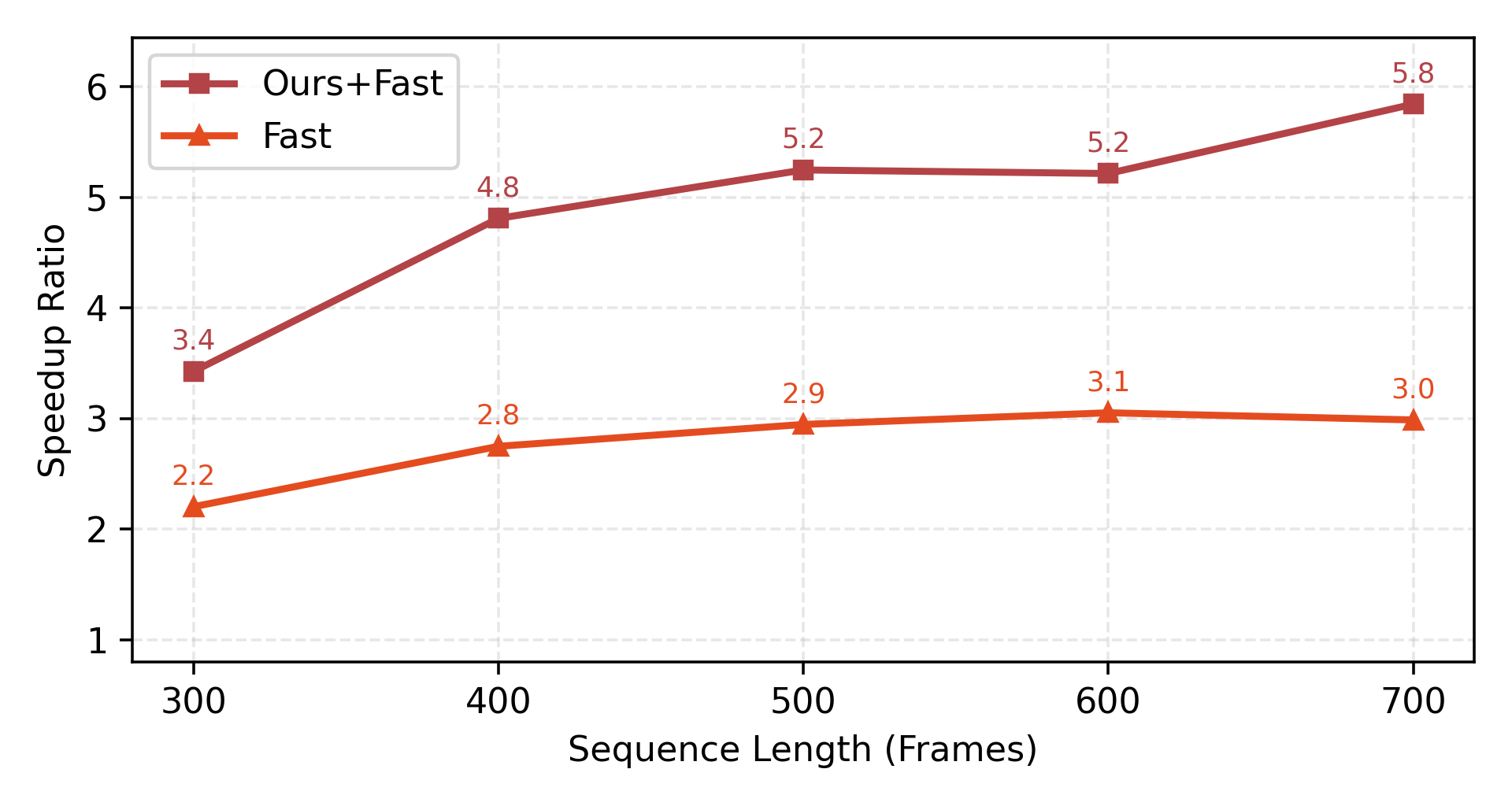}
    \caption{Compounded Speedup (vs. VGGT$^*$) on NRGBD \cite{nrgbd} Results validate the complementarity of frame-level S-VGGT ("Ours") and token-level FastVGGT ("Fast"), showing enhanced acceleration across varying sequence lengths.}
    \label{fig:speed-up-comparison}
\end{figure}

A core claim of our work is that structural redundancy reduction (frame-level) is fully orthogonal to feature redundancy reduction (token-level). To empirically validate this, we evaluate a hybrid configuration, denoted as "Ours+Fast," by combining our frame partitioning module with the token-merging strategy of FastVGGT. As shown in Fig.~\ref{fig:speed-up-comparison}, the hybrid approach significantly improves efficiency compared to FastVGGT alone. For example, "Ours+Fast" achieves a speedup of 3.4$\times$ for a 300-frame sequence, compared to 2.2$\times$ for FastVGGT alone. This trend continues across longer sequences, with "Ours+Fast" reaching up to 5.8$\times$ speedup for a 700-frame sequence, providing a substantial advantage over FastVGGT.

\begin{figure}[htbp]
    \centering
    \includegraphics[width=1.0\linewidth]{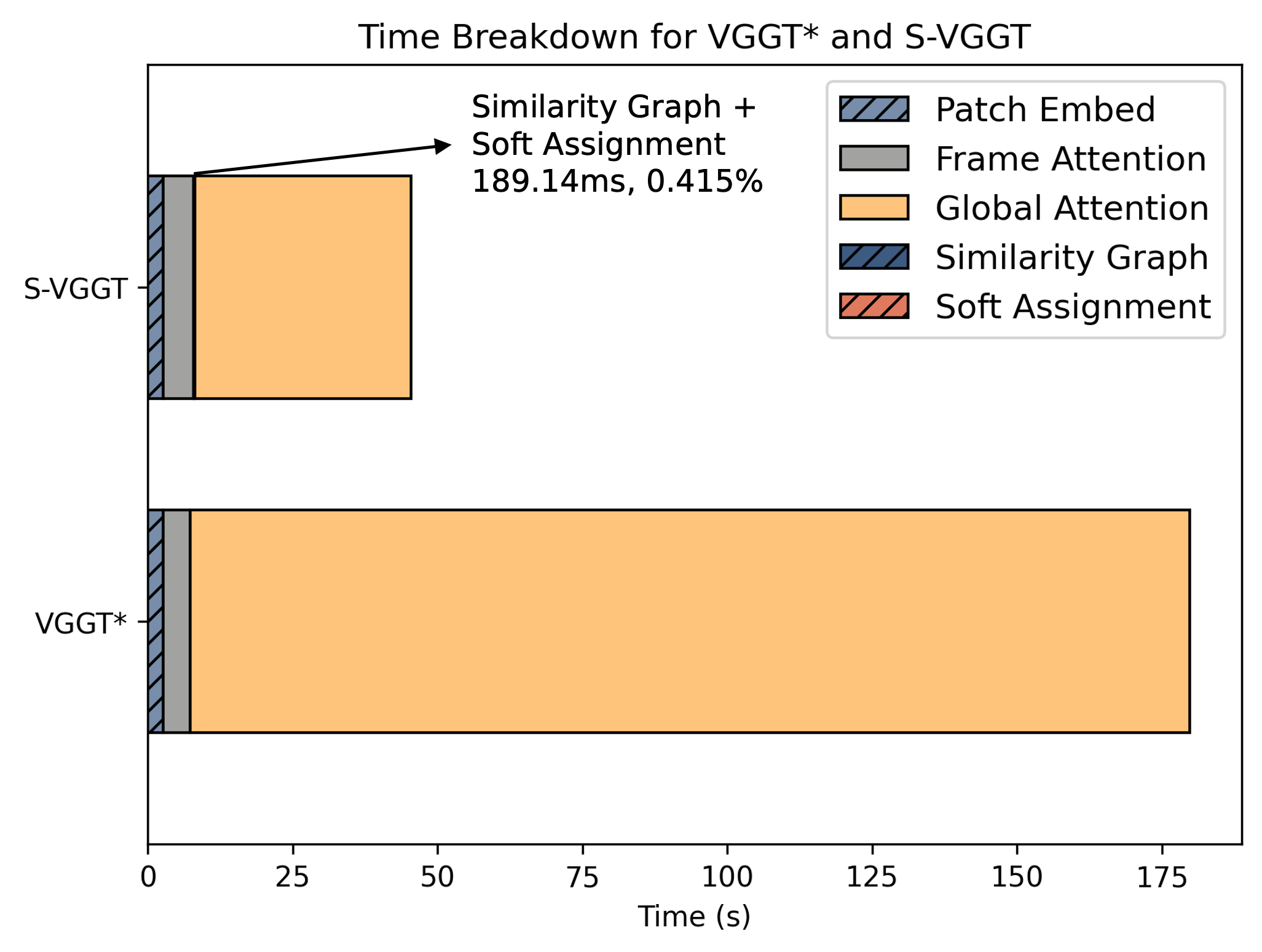}
    \caption{Time breakdown of VGGT$^{*}$ vs. S-VGGT. }
    %
    \label{fig:time_analysis}
\end{figure}

\subsection{Analysis}
While the consistency loss ($\mathcal{L}_{\mathrm{cons}}$) ensures accuracy, the balance ($\mathcal{L}_{\mathrm{bal}}$) and sharpness ($\mathcal{L}_{\mathrm{sharp}}$) terms are critical for computational viability. Their removal leads to unbalanced, degenerate partitions that eliminate the parallel speedup. Furthermore, the Anchor Frame Sharing protocol is essential. Its removal either causes catastrophic geometric misalignment or introduces costly post-hoc optimization, fundamentally compromising our efficient inference goal.

The time breakdown comparison between VGGT* and S-VGGT (Fig.~\ref{fig:time_analysis}) reveals that both methods allocate the majority of their time to global attention. However, S-VGGT introduces two additional modules: similarity graph and soft assignment. While these modules add complexity, they are efficiently integrated, with minimal impact on the overall time cost. This demonstrates that S-VGGT effectively mitigates the global attention bottleneck while maintaining competitive efficiency. The integration of these modules enhances S-VGGT's ability to handle long sequences, making it a scalable solution for 3D perception tasks.

\section{Conclusion}
We address the scalability bottleneck in global attention-based 3D foundation models by introducing S-VGGT, a novel framework that reduces structural redundancy at the frame level. By leveraging density-aware partitioning and anchor frame sharing, our method enables efficient parallel inference. S-VGGT offers significant intrinsic acceleration over the baseline while improving geometric accuracy by mitigating long-range noisy correlations. Additionally, the framework is fully orthogonal to existing token-level techniques, achieving compounded acceleration when combined. This work provides a scalable and efficient solution for high-fidelity 3D perception in future applications.

\bibliographystyle{IEEEbib}
\bibliography{icme2026references}

\end{document}